\documentclass[letterpaper, 10 pt, conference]{ieeeconf}  

\IEEEoverridecommandlockouts                              

\usepackage{graphics} 
\usepackage{epsfig} 
\usepackage{mathptmx} 
\usepackage{times} 
\usepackage{amsmath} 
\usepackage{amssymb}  
\usepackage{cite}
\usepackage{graphicx}
\usepackage{lipsum} 
\usepackage{mathtools}
\usepackage{bm}
\usepackage{cleveref}
\usepackage{epstopdf}
\usepackage{siunitx}
\usepackage{longtable}
\usepackage{booktabs}
\usepackage[utf8]{inputenc}
\usepackage[english]{babel}
\usepackage[dvipsnames]{xcolor}
\usepackage[boxed,lined,commentsnumbered]{algorithm2e}
\usepackage{xcolor}
\usepackage{arydshln}
\usepackage[normalem]{ulem}

\newcommand{\bx}{\mathbf{x}}
\newcommand{\boldf}{\mathbf{f}}
\newcommand{\bX}{\mathbf{X}}

\newcommand{\bQ}{\mathbf{Q}}
\newcommand{\bI}{\mathbf{I}}
\newcommand{\bZ}{\mathbf{Z}}
\newcommand{\bz}{\mathbf{z}}
\newcommand{\boldm}{\mathbf{m}}
\newcommand{\bS}{\mathbf{S}}
\newcommand{\by}{\mathbf{y}}

\newcommand{\bc}{\mathbf{c}}
\newcommand{\bbR}{\mathbb{R}}
\newcommand{\tpose}{\mathsf{T}}
\newcommand{\calN}{\mathcal{N}}
\newcommand{\calD}{\mathcal{D}}
\newcommand{\calO}{\mathcal{O}}

\newcommand{\calX}{\mathcal{X}}
\DeclareMathOperator*{\argmin}{arg\,min}

\title{\LARGE \bf
Wasserstein-Splitting Gaussian Process Regression \\ for Heterogeneous Online Bayesian Inference}

\author{Michael E. Kepler$^{1}$, Alec Koppel$^{2}$, Amrit Singh Bedi$^{2}$, and Daniel J. Stilwell$^{1}$
\thanks{$^{1}$M.E. Kepler and D.J. Stilwell are with the Bradley Department of Electrical and Computer Engineering,
        Virginia Polytechnic Institute and State University, Blacksburg, VA 24060, USA
       {\tt\small mkepler@vt.edu, stilwell@vt.edu}}%
\thanks{$^{2}$A. Koppel and A.S. Bedi are with the Computational and Information Sciences Directorate,
        U.S. Army Research Laboratory, Adelphi, MD 20783, USA 
        {\tt\small alec.e.koppel.civ@mail.mil, amrit0714@gmail.com}}%
\thanks{*This work was supported by the National Defense Science and Engineering Graduate (NDSEG) fellowship program and the Office of Naval Research via grants N00014-18-1-2627 and N00014-19-1-2194}
}

\begin{document}

\maketitle
\thispagestyle{empty}
\pagestyle{empty}

\begin{abstract}
Gaussian processes (GPs) are a well-known nonparametric Bayesian inference technique, but they suffer from scalability problems for large sample sizes, and their performance can degrade for non-stationary or spatially heterogeneous data. In this work, we seek to overcome these issues through (i) employing variational free energy approximations of GPs  operating in tandem with online expectation propagation steps; and (ii) introducing a local splitting step which instantiates a new GP whenever the posterior distribution changes significantly as quantified by the Wasserstein metric over posterior distributions. Over time, then, this yields an ensemble of sparse GPs which may be updated incrementally, and adapts to locality, heterogeneity, and non-stationarity in training data. 
%
%
%
%
We provide a 1-dimensional example to illustrate the motivation behind our approach, and compare the performance of our approach to other Gaussian process methods across various data sets, which often achieves competitive, if not superior predictive performance, relative to other locality-based GP regression methods in which hyperparameters are learned in an online manner. 
\end{abstract}

\section{Introduction}
Gaussian Processes (GPs) are a nonparametric Bayesian inference technique that has been widely used across science and engineering to provide nonlinear interpolation and uncertainty estimates \cite{rasmussen2004gaussian}, as in terrain elevation \cite{vasudevan2009gaussian}, temperature \cite{krause2008near}, the inverse mapping of sensor outputs to joint inputs for the control of a robotic arm \cite{nguyen2009model,deisenroth2013gaussian}, systems identification \cite{pillonetto2014kernel,liu2018gaussiancontrol} and in uncertainty-aware mapping \cite{senanayake2017bayesian,zobeidi2020dense}. In its most essential form, a posterior distribution of an unknown nonlinear map $f:\calX\rightarrow\bbR$ is computed using a basis of feature vectors $\bx_k\in\calX\subset\mathbb{R}^d$, $1\leq k \leq N$, and noisy observations $y_k=f(\bx_k) + \epsilon_k$ where $\epsilon_k$ is observation noise, typically Gaussian. The descriptive power of this framework is hamstrung by two important attributes of the GP posterior: (i) computing the conditional mean and covariance require computational effort cubic $\calO(N^3)$ in the sample size $N$ due to the presence of a kernel matrix inversion; and (ii) consistency guarantees require data to be independent and identically distributed (i.i.d) \cite{van2008rates}. In robotics, data is typically arriving incrementally and exhibits locality or drift \cite{meier2016drifting}. Augmenting GPs to address these issues is the goal of this work.

\begin{table*}[t]
  \centering
  \caption{Comparison of characteristics of WGPR and related works.}
    \begin{tabular}{cccccc}
    \toprule
        Method  & Model representation & Model hyperparameters  & Hyperparameters learned & Splitting criterion & Prediction \\
    \midrule
    LGPR\cite{nguyen2009model}  & Exact & Globally fixed & Offline & Feature space distance & Weighted average \\
    SOLAR GP\cite{wilcox2020solar} & Sparse & Locally adaptive & Online & Feature space distance & Weighted average \\
    WGPR  & Sparse & Locally adaptive & Online & Wasserstein distance & Nearest model \\
    \end{tabular}%
  \label{tab: related work}%
\end{table*}%


Challenge (i), the poor scaling with sample size $N$, originates from the fact that the posterior mean and covariance depend on a data matrix, or kernel dictionary, that accumulates all past observations. A long history of works have studied this issue, and typically select a subset of $M\ll N$ possible model points called inducing \cite{titsias2009variational} or pseudo-inputs \cite{snelson2006sparse}, and optimize GP hyperparameters along this subspace. One may select points via information gain \cite{seeger2003fast}, greedy compression \cite{smola2001sparse}, Nystr{\"o}m sampling \cite{williams2001using}, or other probabilistic criteria \cite{solin2014hilbert,mcintire2016sparse}. A related question is the objective for inducing input optimization, which may be done according to the posterior likelihood along the $M$-dimensional subspace \cite{snelson2006sparse}, or a KL-divergence lower bound thereof called the variational free energy (VFE) \cite{titsias2009variational}. In addition to pseudo-point methods, sparse spectrum Gaussian processes \cite{lazaro2010sparse} can be used to obtain a reduced spectral representation of the Gaussian process. These methods are predominately offline, where the model is trained over the entire training set with a fixed finite number of samples. However, as we discuss next, VFE is more naturally extensible to incremental processing.

In particular, to augment these approaches to apply to settings with incrementally arriving data, dynamic point selection schemes have been proposed \cite{koppel2020consistent}; however, it is difficult to incorporate hyperparameter search, especially for inducing inputs, into these schemes, which are essential to obtaining competitive performance in practice. By contrast, fixed-memory VFE approximations may gracefully incorporate online Expectation Propagation (EP) steps \cite{csato2002gaussian}, as pointed out in \cite{bui2017streaming}, and may be subsumed into the streaming variational Bayes framework \cite{broderick2013streaming}. See \cite{liu2018gaussian,shi2020sparse} for a review of recent advances in Variational Bayes as it pertains to GPs. For this reason, to mitigate the sample complexity bottleneck, we adopt a VFE approach with online EP steps \cite{bui2017streaming}, whose approximate consistency (for i.i.d. settings) is recently established \cite{burt2019rates}. 

Attempts to address issue (ii), i.e., to broaden use of GPs to non-stationary or spatially heterogeneous data, have taken inspiration from vector-valued time-series analysis as well as ensemble methods. Specifically, time-series approaches seek to define covariance kernels that vary with time and/or space  \cite{wilson2015kernel,remes2017non,toth2019variational, paciorek2003nonstationary}. These approaches have also given rise to use of GPs in high-dimensional problems through use of convolutional kernels \cite{blomqvist2019deep}. However, doing so typically requires offline training, and numerical conditioning issues make it challenging to adapt them to the online setting, which is inherent to continually learning autonomous robots. Efforts to address the non-stationarity that arises in problems with dynamics have also been considered, especially in the context of occupancy grids for mapping application \cite{senanayake2018automorphing,senanayake2017bayesian}.

Our approach is, instead, inspired by {ensemble} methods, where one builds a family of ``weak learners" and forms the inference by appropriately weighting the constituents \cite{littlestone1994weighted,fudenberg1995consistency}, as in boosting \cite{freund1997decision}, and local regression, i.e., the Nadaraya-Watson estimator \cite{nadaraya1964estimating,watson1964smooth,vijayakumar2005incremental}. Two design decisions are crucial to formulating an ensemble GP: (a) how to form the aggregate estimate, and (b) when to instantiate a new model.
The line of research related to ensemble GPs began with local Gaussian Process regression (LGPR) \cite{nguyen2009model}, which (a) formulates estimates by weighted majority voting akin to Nadaraya-Watson, and (b) assigns samples to a local model based on a similarity measure defined in terms of the covariance kernel to the arithmetic mean of a local model's samples. If no local models are sufficiently close, then a new local model is instantiated. We note that this approach requires hyperparameters to be fixed \emph{in advance} of training. Most similar to this work is sparse online locally adaptive regression using Gaussian processes (SOLAR GP) \cite{wilcox2020solar}, where each local GP is approximated using VFE and updated online using EP \cite{bui2017streaming}. 

The aforementioned approaches to ensemble GPs have two key limitations \cite{nguyen2009model,meier2014incremental,wilcox2020solar}: (a) they form aggregate estimates by weighted voting, and (b) use distance in feature space to quantify whether a new constituent model is needed. Weighted voting, as inspired by local regression, hinges upon the validity of an i.i.d. hypothesis \cite{nadaraya1964estimating,watson1964smooth}. Moreover, using distance in feature space may falsely ascribe novelty to far away samples that are similar in distribution to a current local model. To address these issues, we propose Wasserstein-splitting Gaussian Process Regression (WGPR), whose main merits are as follows:
\begin{itemize}
\item the notion of similarity is specified as the Wasserstein distance, a rigorous metric over distributions which is available in closed-form for Gaussian Processes. This distance is used to quantify the change in the each model's predictive distribution caused by an update with the new training data, and hence defines a decision rule for when to instantiate a new local model: a new local model is instantiated only if the change in the posterior predictive model exceeds a threshold in the Wasserstein distance for every model, otherwise it is incorporated into the most similar, i.e. smallest Wasserstein distance, model via expectation propagation (EP) steps;

%
\item WGPR forms predictive distributions upon the basis of only the the nearest local model in terms of Euclidean distance in the input space, rather than a weighted vote, in order to more strictly encapsulate local heterogeneity in the data, and hence only operates upon a \emph{local} i.i.d. hypotheses;
\item experimentally WGPR identifies the number of data phase transitions on a 1-dimensional data set and achieves competitive performance with the state of the art for a variety of real data sets.
\end{itemize}
We compare characteristics of WGPR, LGPR\cite{nguyen2009model}, and SOLAR GP\cite{wilcox2020solar} in Table \ref{tab: related work}. Note that LGPR requires the hyperparameters of all models to be fixed to the same set of global hyperparameters, which are learned \textit{a priori} offline. WGPR, on the other hand, does not require the hyperparameters to be globally fixed \textit{a priori}, but allows for the hyperparameters of a model to be updated recursively as more data is collected. In contrast to both LGPR and SOLAR GP, WGPR decides whether to update an existing model or instantiate a new one based on the effect the new data has on the posterior distribution of each model, quantified in terms of the Wasserstein distance, and prediction is based on the model nearest to the query point in an effort to preserve local features, which can be filtered out when using a weighted prediction of models.

\section{Background and Notation}\label{sec: BG}
In WGPR, we employ a sparse approximation of the posterior predictive distribution for each model. In this section, we introduce notation and provide the background material on sparse Gaussian processes required to implement WGPR.
\subsection{Gaussian Process Regression}
A Gaussian process is a collection of random variables, any finite number of which have a joint Gaussian distribution \cite{williams2006gaussian}. We use a Gaussian process to model the spatial field of interest $f:\calX\rightarrow\bbR$, which we write as $f(\cdot) \sim GP\left(m(\cdot),k(\cdot,\cdot)\right)$, where $m:\calX \rightarrow \bbR$ and $k:\calX \times \calX \rightarrow \bbR$ denote the mean and covariance functions, respectively. Practical application involves evaluating a Gaussian process at a finite set of inputs $\bX =\{\bx_1,\cdots,\bx_N\}\subset \calX $, and with a slight abuse of notation we write $f(\bX) \sim \mathcal{N}\left(m(\bX),k(\bX,\bX)\right)$
where the mean vector $m(\bX)\in \bbR^N$ and covariance matrix $k(\bX,\bX)\in \bbR^{N \times N}$ are defined element-wise via $[m(\bX)]_i = m(\bx_i)$ and 
$[k(\bX,\bX)]_{i,j} = k(\bx_i,\bx_j)$. More generally, for $\bZ =\{\bz_1,\cdots,\bz_M\}\subset \calX $, the matrix $k(\bX,\bZ)\in \bbR^{N \times M}$ is defined by $[k(\bX,\bZ)]_{i,j} = k(\bx_i,\bz_j)$ for all $i=1:N$, and all $j=1:M$.

Gaussian process regression follows a Bayesian approach which begins with specifying a prior distribution $f(\cdot) \sim GP_0\left(m_0(\cdot),k_0(\cdot,\cdot)\right)$.
As is customarily done for the sake notational simplicity \cite[Ch. 2.2]{williams2006gaussian}, we select the zero-mean function as the prior mean function $m_0(\cdot)$. The covariance kernel $k_0(\cdot,\cdot)$ is user-specified and encodes prior assumptions, such as smoothness, about the underlying function $f$. Throughout this paper, we employ the commonly-used squared exponential covariance kernel given by
\begin{align}\label{eq: SE Kernel}
    k_0(\bx,\bx')=\sigma_f^2 \exp\{-({1}/{2})(\bx-\bx')^{\tpose}\Lambda^{-1}(\bx-\bx')\},
\end{align}
where \(\sigma_f^2\) denotes the signal variance, and \(\Lambda = \textrm{diag}(\lambda_1^2,...,\lambda_d^2)\) contains the length-scale parameter $\lambda_i$ of each input dimension $i=1,...,d$. The entire conceptual development proceeds analogous for other covariance kernels. The assumption that each measurement is corrupted by independent identically distributed Gaussian noise $\epsilon \sim N(0,\sigma_n^2)$ gives rise to a Gaussian likelihood, and given a collection of training data $\{(\bx_i,y_i)\}_{i=1}^N$ the posterior process is
\begin{align}\label{eq: post}
    f(\bx)\sim GP(m_{post}(\bx), k_{post}(\bx,\bx')),
\end{align}
where
\begin{align*}
    m_{post}(\bx) &= k_0(\bx, \bX)(k_0(\bX,\bX)+\sigma_n^2\mathbf{I})^{-1}\by, \\
    k_{post}(\bx,\bx') &= k_0(\bx, \bx') + k_0(\bx, \bX)(k_0(\bX,\bX)+\sigma_n^2\mathbf{I})^{-1}k_0(\bX, \bx'),
\end{align*}
for all $\bx,\bx'\in \calX$. Performing inference with the exact form of the posterior \eqref{eq: post}, is only feasible for small to moderate data sets, e.g. sample size $N<10,000$, as the necessary matrix inversion in the preceding expression for the posterior mean and covariance functions incurs a computational cost that scales $\calO(N^3)$. To mitigate the complexity bottleneck in the sample size, we employ a sparse approximation of the posterior distribution for each Gaussian process parameterization henceforth considered. Next we define the specific sparse approximation scheme we consider.

\subsection{Variational Free Energy Approximation} \label{ss: VFE sparse}
We reduce computational and memory costs using the posterior approximation developed by Titsias \cite{titsias2009variational}. The key idea underpinning the approximation is to express the posterior in terms of $M \ll N$ pseudo (inducing) points, which serve to summarize the exact posterior associated with $N$ training points. With $M$ fixed (by the user), the pseudo points $\bZ = \{\bz_1,...,\bz_M\}\subset \calX$, specified as variational parameters, are selected so as to maximize a lower bound on the true log marginal likelihood. This bound is referred to as the variational free energy and is given by
\begin{align} \label{eq: VFE}
    F_{VFE} = \log[\mathcal{N}(\by;\mathbf{0},\sigma_n^2\bI + \bQ_{NN})]-\frac{1}{2\sigma_n^2}\mathrm{tr}(k_0(\bX,\bX)-\bQ_{NN}),  
\end{align}
where $\bQ_{NN} = k_0(\bX,\bZ)k_0(\bZ,\bZ)^{-1}k_0(\bZ,\bX)$. Equivalently, by maximizing the variational free energy, the Kullback-Leibler divergence between the variational distribution and exact posterior distribution over $f(\bZ)$ is minimized. Given the pseudo inputs $\bZ$, the optimal variational distribution over $f(\bZ)\triangleq \boldf_Z$ has an analytic form $\boldf_Z \sim N(\mu_Z,\bS_Z)$ with
\begin{align*}
    \mu_Z &= \sigma_n^{-2}k_0(\bZ,\bZ)\Sigma k_0(\bZ,\bX)\by, \\
    \bS_Z &= k_0(\bZ,\bZ)\underbrace{(k_0(\bZ,\bz)+\sigma_n^{-2} k_0(\bZ,\bX)k_0(\bX,\bZ))^{-1}}_{\triangleq \Sigma}k_0(\bZ,\bZ).
\end{align*}
Given the optimal pseudo-inputs $\bZ\subset \calX$, the approximate posterior is given by $f(\cdot)\sim GP(m_{VFE}(\cdot),k_{VFE}(\cdot,\cdot))$, where
\begin{align} \label{eq: vfe post}
    m_{VFE}(\bx) = &k_0(\bx,\bZ)k_0(\bZ,\bZ)^{-1}\mu_{Z}, \\
    k_{VFE}(\bx,\bx') =& k_0(\bx,\bx') - k_0(\bx,\bZ)k_0(\bZ,\bZ)^{-1}k_0(\bZ,\bx') \nonumber\\
                      &+ k_0(\bx,\bZ)k_0(\bZ,\bZ)^{-1}\bS_{Z}k_0(\bZ,\bZ)^{-1}k_0(\bZ,\bx').\nonumber
\end{align}
In practice, we first obtain the pseudo-inputs $\bZ \subset \calX$ and hyperparameters $\theta = \{\sigma_f,\Lambda, \sigma_n\}$ via gradient-based minimization of the negative variational free energy \eqref{eq: VFE}. With the (locally optimal)\footnote{Observe that in general, since the VFE is non-convex with respect to its hyperparameters $\theta$ and pseudo-inputs $\bZ$, the best pointwise limit one may hope to achieve via gradient-based search is a local minimizer.}  hyperparameters and pseudo-inputs in hand, we then form the posterior predictive distribution \eqref{eq: vfe post}. 

 As new training data is collected, we use the principled \emph{expectation propagation} framework of Bui et al.\cite{bui2017streaming} to recursively update the existing sparse approximation, along with the hyperparameters and pseudo points. For the recursive update, it is assumed that previous measurements are inaccessible so knowledge about previously collected measurements is inferred through existing pseudo points. More precisely, let $GP_{old}$ denote the current sparse posterior approximation with associated pseudo inputs $\bZ_a \subset \calX$ and hyperparameters $\theta_{old}$. The updated posterior approximation follows from an optimization problem, where given the new data $\calD_{new} = \{(\bx_i,y_i)\}_{i=1}^{N_{new}}$ and $\bZ_a \subset \calX$ we maximize the online log marginal likelihood $F_{OVFE}$. The optimization gives rise to the updated posterior predictive distribution
\begin{align} \label{eq: OVFE post}
    GP_{upd}(m_{upd}(\cdot), k_{upd}(\cdot,\cdot)),
\end{align}
which depends on the new pseudo inputs $\bZ_b$ and hyperparameters $\theta_{new}$ that jointly optimize $F_{OVFE}$. We omit the exact expressions for $F_{OVFE}$ and the posterior mean and covariance as they are prohibitively long, but they can be readily accessed from the supplementary appendix of \cite{bui2017streaming}. In summary, the recursive update is performed by first obtaining the pseudo-inputs $\bZ_b$ and hyperparameters $\theta_{new}$ that optimize $F_{OVFE}$, and then using them to form the posterior \eqref{eq: OVFE post}.

\section{Wasserstein-Splitting Gaussian Processes}\label{sec: GP}
In this section, we present Wasserstein-Splitting Gaussian Process Regression (WGPR). The high-level structure of WGPR mirrors that of LGPR \cite{nguyen2009model} and SOLAR GP \cite{wilcox2020solar}, however our method deviates from the existing methods in two notable ways: (i) we instantiate new models on the basis of the their posterior predictive distribution, rather than their locality in the input space, and (ii) we propose an alternative approach for making predictions.

\subsection{Training}\label{subsec:training}
{\bf \noindent Initialization:} Training the overall prediction model begins with an empty collection of models, and given the initial batch of training data $\calD_{new} =\{(\bx_i,y_i)\}_{i=1}^{N_{new}}$ of $N_{new}$ samples, we instantiate the first local model using the VFE approximation as detailed in Section \ref{ss: VFE sparse} yielding $GP_1$.

{\bf \noindent Updating the Current Ensemble:} Now, suppose we have a collection $\{GP_{j}\}_{j=1}^J$ of $J$ Gaussian Process models. Upon obtaining a new batch of training data $\calD_{new}$, the key question is how to decide which model is most similar, and thus is a suitable candidate for executing an update. 
Previous approaches\cite{vijayakumar2005incremental,nguyen2009model,wilcox2020solar} define similarity in terms of the covariance kernel [cf. \eqref{eq: SE Kernel}]. With this definition of similarity, two distant regions of the spatial field, characterized by the same set of underlying hyperparameters, gives rise to two different models. There are indeed apparent structural similarities to local kernel regression estimates of conditional expectations such as the Nadaraya-Watson estimator \cite{nadaraya1964estimating,watson1964smooth} or ensemble techniques \cite{freund1997decision} inspired by majority voting  \cite{littlestone1994weighted,fudenberg1995consistency}. However, the former approach hinges upon i.i.d. training examples from a stationary distribution, and the latter collapses distributional inference to point estimates in feature space, which belies the fact that we are in the Bayesian setting. 

%
 %
%
We, instead, base similarity on the intuition that if $\calD_{new}$ is similar to an existing model, then the new data should minimally change the posterior predictive distribution according to some metric over probability distributions. In other words, predictions of the updated individual model (i) remain (approximately) unchanged across previously seen data, and (ii) agree with the predictions of a newly instantiated model across the new measurements. 
To formalize this intuition and assess how the new data would affect each posterior distribution in the collection $\{GP_j(m_j(\cdot), k_j(\cdot,\cdot))\}_{j=1}^J$, we update each model using the recursive VFE framework \cite{bui2017streaming} using the new mini-batch $\calD_{new}$, as outlined in Section \ref{sec: BG}, which yields the updated collection $\{GP_j^{upd}\}_{j=1}^J$ where
\begin{align*}
    GP_{j}^{upd} \left(m_{j}^{upd}(\cdot), k_{j}^{upd}(\cdot,\cdot)\right)\; , \quad j=1,...,J.
\end{align*}
 Let $\bZ_j \subset \cal X$ denote the pseudo-inputs of model $j$, prior to performing the update with $\calD_{new}$. We then measure the change in predictions over previous data by computing the squared Wasserstein distance \cite{panaretos2019statistical} between the original and updated posterior evaluated at $\bZ_j$, i.e.  
\begin{align*}
    w_j^{old} =& \Vert \boldm_{j} \!- \!\boldm_{j,Z}^{upd} \Vert^2\!+\! \textrm{tr}\left(\bS_j \!+\! \bS_{j,Z}^{upd} -2\left(\bS_j^{1/2}\bS_{j,Z}^{upd}\bS_j^{1/2}\right)^{1/2}\right),
\end{align*}
where $\Vert \cdot \Vert$ denotes the Euclidean norm, and
\begin{align*}
    \boldm_j &\triangleq m_{j}(\bZ_j), & \bS_j &\triangleq k_{j}(\bZ_j,\bZ_j) \\
    \boldm_{j,Z}^{upd} &\triangleq m_{j}^{upd}(\bZ_j), & \bS_{j,Z}^{upd} &\triangleq k_{j}^{upd}(\bZ_j,\bZ_j).
\end{align*}
Note that the Wasserstein distance is a valid metric over probability measures, and its salient feature is that it is computable in closed-form for Gaussians, as well as it avoids some of the computational issues of comparable choices such as the Total Variation or Hellinger metrics -- see \cite{wasserman2006all}. 

 Now, with $\{GP_{j}^{upd}\}_{j=1}^J$ in hand, we quantify how well each updated model agrees with the new data. To do so, we take $\calD_{new}$ and instantiate a new approximate GP via the batch variational free energy sparse approximation \eqref{eq: VFE} - \eqref{eq: vfe post}:
\begin{align*}
    GP_{J+1} \left(m_{J+1}(\cdot), k_{J+1}(\cdot,\cdot)\right)
\end{align*}
that is free from bias of any previous data. Ideally, if model $j$ is similar to the new data, then the distributions $GP_{j}^{upd}$ and $GP_{J+1}$ should agree well over the new data.  Let $\bX_{new} \subset \cal X$ denote the locations of the new measurements. We measure the agreement by computing the squared Wasserstein distance between the newly instantiated and updated posterior evaluated at $\bX_{new}$, given by
\begin{align*}
    w_j^{new} \!\!\!=\!\! \Vert \boldm_{J\!+\!1} \!-\! \boldm_{j,X}^{upd} \Vert^2\!\!+\! \textrm{tr}\left(\!\!\bS_{J\!+\!1} \!\!+\! \bS_{j,X}^{upd} \!-\!2\left(\bS_{J+1}^{1/2}\bS_{j,X}^{upd}\bS_{J\!+\!1}^{1/2}\right)^{\!\!1/2}\right),
\end{align*}
where
\begin{align*}
    \boldm_{J+1} &\triangleq m_{J+1}(\bX_{new}), & \bS_{J+1} &\triangleq k_{J+1}(\bX_{new},\bX_{new}) \\
    \boldm_{j,X}^{upd} &\triangleq m_{j}^{upd}(\bX_{new}), & \bS_{j,X}^{upd} &\triangleq k_{j}^{upd}(\bX_{new},\bX_{new}).
\end{align*}
Thus, the net measure of similarity $w_j$ is the aggregation of the squared-Wasserstein distance between the previous model $GP_{j}$ and its updated variant $GP_{j}^{upd}$ evaluated at the old pseudo-inputs $\bZ_j$, which serve as a proxy for previously seen data, and the squared distance between $GP_{j}^{upd}$ and the new model $GP_{J+1}$ evaluated at the new measurement locations $\bX_{new}$:
\begin{align} \label{eq: sim metric}
    w_j = w_j^{new} + w_j^{old},   & & j =1,\dots,J.
\end{align}
%
%
\begin{algorithm}[t]
\small
\SetKwInOut{Input}{Input}\SetKwInOut{Output}{Output}
\SetKwInOut{Input}{Input}\SetKwInOut{Output}{Output}
\Input{Instantiation threshold $\epsilon$}
\Output{Collection of models $\{GP\}_{j=1}^{J_{k+1}}$}
\BlankLine
$\calD_{new}^1 \leftarrow$ Receive $1^{st}$ batch of data\;
$GP_{1} \leftarrow$ Init. new $GP$ with $\calD_{new}^1$  (Eq. \eqref{eq: vfe post})\; 
\For{k = 2,3,...}{
$\calD_{new}^k \leftarrow$ Receive new batch of data\;
$GP_{J_k+1} \leftarrow$ Init. new $GP$ with $\calD_{new}^k$  (Eq. \eqref{eq: vfe post})\; 
\For{$j = 1 : J_k$}{ 
    $GP_j^{upd}\leftarrow$ Update $GP_j$ with $\calD_{new}^k$  (Eq. \eqref{eq: OVFE post})\;
    $w_j^{old}\leftarrow$ Similarity wrt previous predictions\;
    $w_j^{new}\leftarrow$ Similarity wrt new data\;
    $w_j = w_j^{old} + w_j^{new}$ \; 
}
$w_{j^*} = \min{w_j}$ \;
\eIf{$w_{j^*} \leq \epsilon$}{
  $\{GP_{j}\}_{j=1}^{J_{k+1}} \leftarrow   \{GP_{j}\}_{j=1}^{J_k}\setminus  \{GP_{j*}\} \cup  \{GP_{j*}^{upd}\}$\;
  }{
  $\{GP_{j}\}_{j=1}^{J_{k+1}} \leftarrow    \{GP_{j}\}_{j=1}^{J_k}\cup  \{GP_{J_k+1}\}$.
  }
}
\vspace{0.5mm}
\caption{Wasserstein-Splitting GP Regression (WGPR): train online  approximate GP ensemble.
} \label{alg: upd collection}
\normalsize
\end{algorithm}
Once we compute $w_j$ for all $j$, we define the most similar model $GP_{j^*}$ as the one that minimizes \eqref{eq: sim metric} , i.e., $j^* = \argmin_{j=1,\dots,J}{w_j}$ and provided it is sufficiently similar, meaning $w_{j^*}$ is less than a user-defined threshold $\epsilon$, we retain the updated model $GP_{j*}^{upd}$. All other models revert back to their previous state, prior to the update, and the newly instantiated model $GP_{J+1}$ is discarded. However, if $GP_*$ is not sufficiently similar, then we retain $GP_{J+1}$ and revert all local models back to their state prior to the update. 

To be more precise, denote $J_k$ as the number of ensemble models at time $k$. Upon observing mini-batch $\calD_{new}^k$, we update model $GP_{j^*}$ if \eqref{eq: sim metric} is less than a threshold $\epsilon$, otherwise we add $GP_{J_K + 1}$ to the collection; succinctly stated as
\begin{align} \label{eq:instantiation}
   \{GP_{j}\}_{j=1}^{J_{k+1}} &\leftarrow   \left\{\{GP_{j}\}_{j=1}^{J_k}\setminus  \{GP_{j*}\}\right\}  \cup  \{GP_{j*}^{upd}\} &&\text{ if } w_{j^*} \leq \epsilon \nonumber \\
   \{GP_{j}\}_{j=1}^{J_{k+1}} &\leftarrow    \{GP_{j}\}_{j=1}^{J_k}\cup  \{GP_{J_k+1}\}&& \text{ if } w_{j^*}> \epsilon
\end{align}
We summarize the process of iteratively developing the collection of models $\{GP_{j}\}_{j=1}^{J_k} $ in Algorithm \ref{alg: upd collection}. 

\subsection{Model Evaluation and Prediction}\label{subsec:prediction}
The previous approaches \cite{vijayakumar2005incremental,nguyen2009model,wilcox2020solar} propose combining predictions from various models to form a single weighted prediction, as mentioned in the preceding section, mostly inspired by ensemble \cite{freund1997decision} and local kernel smoothing techniques \cite{nadaraya1964estimating,watson1964smooth}. However, the validity of doing so hinges upon an i.i.d. stationarity hypothesis for the unknown data distribution. This makes these approaches well-suited to capture global patterns (long-term spatial correlations) in data at the expense of filtering out local patterns \cite{liu2018gaussian}. In an effort to preserve the non-stationary features of a heterogeneous spatial field, we predict the spatial field at a location of interest $\bx_s$ by considering the model with the pseudo-input closest to $\bx_s$ evaluated as 
%
$GP_* = \argmin_{j=1:J}{\left\{\min_{\bz_i\in\bZ_j}{\Vert \bz_i - \bx_s \Vert^2}\right\}}$,
and using $GP_*(m_*(\cdot), k_*(\cdot,\cdot))$ to predict $f(\bx_s)\sim \calN(\mu_s, \sigma_s^2)$ where
    $\mu_s = m_*(\bx_s)$ and  $\sigma^2_s = k_*(\bx_s,\bx_s)$,
%
with the posterior given by the VFE approximation \eqref{eq: vfe post} for models that have not undergone a recursive update. Models that have been recursively updated have the posterior mean and covariance functions given by \eqref{eq: OVFE post}. This better encompasses non-stationarity and heterogeneous nonlinearities, although it may be susceptible to sub-sampling bias or discontinuities as a respective function of the the number of pseudo-inputs in a local model or the kernel hyperparameters.


\begin{figure}
  \centering
  \includegraphics[width = \linewidth,height=0.28\textheight]{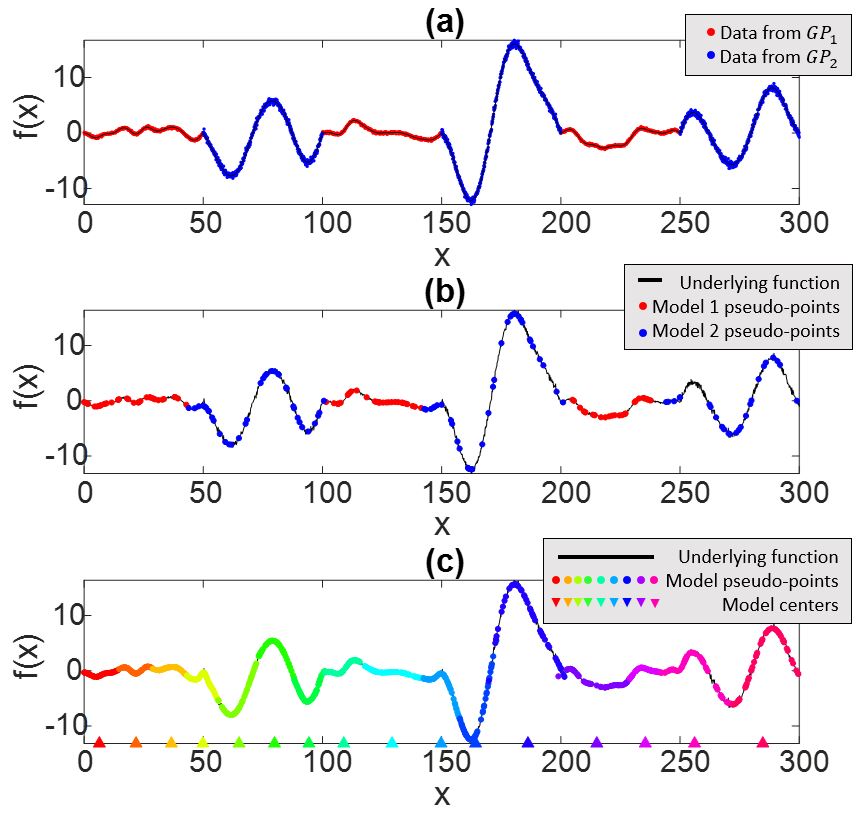}
  \caption{Toy example motivating our definition of similarity. The training data in (a) is generated from two distinct Gaussian processes with known hyperparameters. By defining similarity in terms of the posterior predictive model, WGPR identifies the two models as seen in (b). Other methods that define similarity based on proximity in input space fail to capture to the two underlying models succinctly as indicated by the 16 models in (c).\vspace{-7mm}}
\label{fig: toy ex}
\end{figure}

\section{Empirical Assessment}\label{sec: Results}
In this section, we explore the performance of WGPR across various numerical experiments. We start with a 1-dimensional toy example that motivates defining model similarity in terms of the Wasserstein metric over posterior distributions rather than simply distance in feature space (or a positive definite map thereof). Next, the performance of WGPR is compared to SOLAR-GP \cite{wilcox2020solar}, which has the same algorithmic structure and Gaussian process model representation, but defines similarity in terms of Euclidean distance. The purpose of this comparison is to highlight the difference in performance due to the new similarity metric and using the prediction of the nearest model. Lastly, we compare the performance of several methods, that learn hyperparameters online, with LGPR, which is optimal in the sense that hyperparameters are learned offline on a training set characteristic of the entire domain. In all comparisons, we evaluate the methods on a real-world bathymetry dataset collected at Claytor Lake in Southwest Virginia, as well as four other publicly accessible datasets\footnote{\noindent Kin40k: https://www.cs.toronto.edu/$\sim$delve/data/kin/desc.html \\ Abalone: https://archive.ics.uci.edu/ml/datasets/abalone\\ Sarcos Joint 1: http://www.gaussianprocess.org/gpml/data/ \\ Pumadyn(8nm): https://www.cs.toronto.edu/$\sim$delve/data/pumadyn/desc.html}. Our MATLAB-based implementation uses the gradient optimizer and covariance kernel functionality provided by the GPML toolbox\cite{rasmussen2010gaussian}.

\subsection{Toy-example} \label{ss: toy}
To illustrate the motivation behind our definition of similarity, consider the non-stationary dataset depicted in Figure \ref{fig: toy ex}(a). This dataset was constructed from two different Gaussian process models with known hyperparameters. Ideally, this data set could be concisely described with two models. The dataset is processed in mini-batches of size $N_{new} = 100$ in spatial order from $\bx = 0$ to $\bx = 300$ and each local model employs $M=50$ pseudo points. Provided the instantiation threshold $\epsilon$ is set appropriately, WGPR is able to identify the two underlying models as seen in \ref{fig: toy ex}(b). Regardless of how the instantiation threshold is set, SOLAR GP is incapable of succinctly identifying the two underlying models in \ref{fig: toy ex}(a) because a new model is instantiated provided new data is sufficiently far from all other existing models. This phenomenon is demonstrated by the 16 models in \ref{fig: toy ex}(c). WGPR, with 2 models, achieves a root mean squared error (RMSE) of 0.26. SOLAR GP, with 16 models, achieves an RMSE of 0.39.

\subsection{Comparison of Online Local GPs on Real Data} \label{ss: online comp}
To examine how our new definition of similarity and making predictions based on the nearest model impact performance, we conducted experiments to examine the difference in performance between WGPR and SOLAR GP \cite{wilcox2020solar}.  In theory, the model instantiation threshold can be set to achieve anywhere from a single model to the maximum allowable number of models (as dictated by the number of batches of training data). So ideally, the comparison should highlight the performance difference between the two methods as a function of the number of models used. Thus, the instantiation thresholds are set separately for WGPR and SOLAR to yield approximately the same number of models. Training data was processed in batches of size $N_{new}=100$ and we employed $M=50$ pseudo points for each individual Gaussian process model.

Additionally, we note that for WGPR, computing the similarity metric $w_j$ for each $j=1:J$ can be computationally prohibitive. In practice, we can reduce computational complexity by computing the similarity metric over the $\hat{J}\leq J$ models closest in terms of Euclidean distance in input space given by $\Vert \bc_j - \bc_{new} \Vert ^2$ for $j =1,\dots,J$, where $\bc_{j}\in \calX$ is the arithmetic center of the pseudo-inputs of model $j$ and $\bc_{new}\in \calX$ is the arithmetic center of the new batch of data. For the experiments in this subsection and the following, we fix $\hat{J}=5$. Although the $\hat{J}$ nearest models are selected on the basis of proximity, similar to \cite{nguyen2009model,wilcox2020solar}, the decision to instantiate a new model or update an existing one is ultimately dictated by the similarity measure $\eqref{eq: sim metric}$. This distinction is particularly important for cases in which two regions, that are close in proximity, are characterized by different spatial fields. Provided the regions are sufficiently close, a distance-based similarity metric will incorporate these two different spatial fields into one. In contrast, we can use the similarity measure \eqref{eq: sim metric}, based on predictive distributions, to identify that these two regions are best-characterized by two models and thereby employ two models to avoid the pathology.

For the comparison of SOLAR GP and WGPR, we began by setting the instantiation thresholds to yield the maximum number of allowable models and progressively adjusted the thresholds to yield fewer models. For SOLAR GP, despite meticulous tuning of the number of pseudo points and mini-batch size, we regularly observed divergence in the Claytor lake, Abalone, and Sarcos data sets. By divergence, we mean that the posterior mean evaluated at the pseudo-inputs, i.e $m(\bZ_b)$, tended to infinity for some local models. This is due to the online update of the VFE posterior approximation \cite{bui2017streaming}, which only guarantees we recover the batch posterior and marginal likelihood approximation \cite{titsias2009variational} when the hyperparameters and pseudo-inputs are fixed in advance. When operating with online updates, however, it is possible that pseudo-inputs become very concentrated in a small region of the feature space, causing kernel matrices to become near-singular.
\begin{figure}
  \centering
  \includegraphics[width = \linewidth]{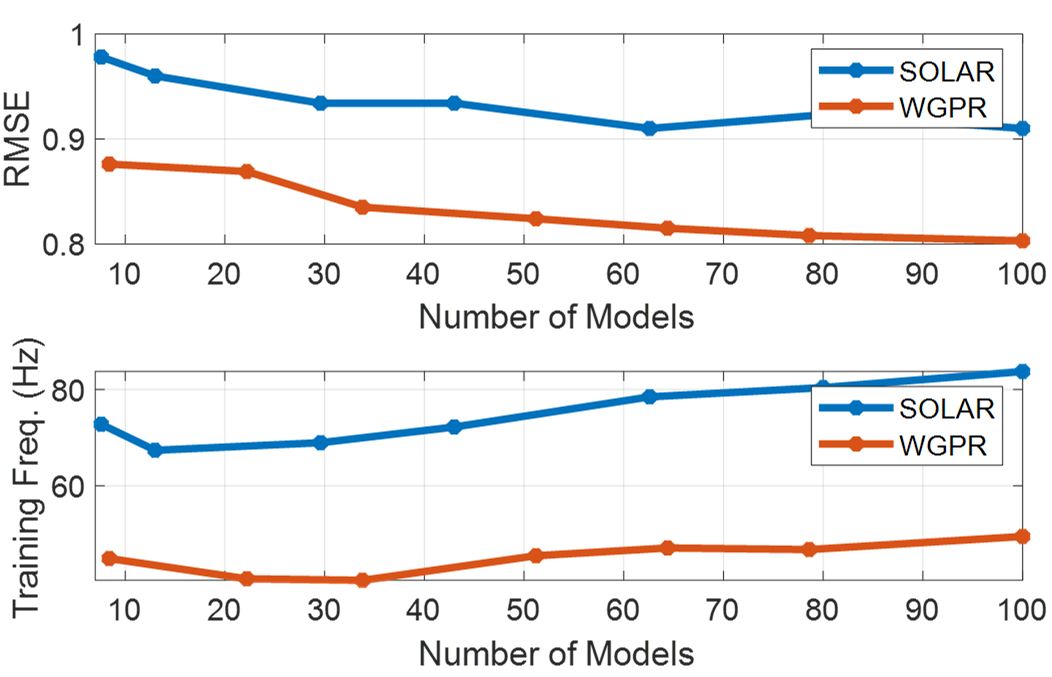}
  \caption{Comparison of the performance difference between our method (WGPR) and SOLAR GP \cite{wilcox2020solar} on the kin40k dataset. Regardless of the number of models used, WGPR outperforms the other online methods at the expense of additional computations required to compute new similarity metric.}
\label{fig: solar comp}
\end{figure}

With our definition of similarity \eqref{eq: sim metric}, on the other hand, we can identify when a stable update can be performed, as the effect of the posterior update is reflected in our similarity metric. 
In other words, if an EP update with the new data $D_{new}$ causes numerical instability for model $j$, then clearly model $j$ is not a suitable candidate to be updated. For our numerical experiments with WGPR, we observed this had the effect of preserving convergence, and overall requiring significantly fewer models on a consistent basis than SOLAR GP. For example, given the batch size of $N_{new}=100$, the maximum number of models that can be instantiated with the Claytor Lake dataset is 127. We were consistently able to achieve as few as 14.6 models, which is the average across 5 trials, with standardized MSE of 0.097, whereas SOLAR GP only achieves sensible predictive performance when there are more than 108.8 models, on average, with standarized MSE of 0.183. Thus, for the Claytor Lake, Sarcos, and Abalone data sets, we can only non-trivially compare predictive performance between SOLAR GP and WGPR in the setting where a new model is instantiated with each new batch of data. We observed that WGPR outperformed SOLAR on the Claytor Lake and Sarcos data sets, and both methods achieved approximately the same predictiableve performance on the Abalone data set.  

For the Kin40k dataset, however, we were able to achieve stable performance for SOLAR GP across various numbers of models employed. We illustrate the overall performance difference between WGPR and SOLAR GP in Figure \ref{fig: solar comp}. We report the root MSE, along with the training frequency achieved by both methods. Training frequency refers to how many samples are processed per second. Observe that for a given number of models, WGPR outperforms SOLAR GP in terms of predictive performance. This is at the expense of additional computational complexity, as reflected by the lower training frequency. Additionally, we note that the predictive performance of approximately 10 models with WGPR is superior to SOLAR GP's best-case predictive performance which occurs with approximately 100 models. For applications in which the size of the map representation is critical, as in communication-limited applications, this is a significant advantage for WGPR.

We conclude by stating that we were also able to achieve stable performance for SOLAR GP on the Pumadyn dataset and observed that SOLAR GP (RMSE = 1.088) outperformed WGPR (RMSE = 1.142). However, performance was approximately constant across the number of models used, and thus we omit the performance plots for the sake of space limitations.  

\begin{figure}
  \centering
  \includegraphics[width = \linewidth]{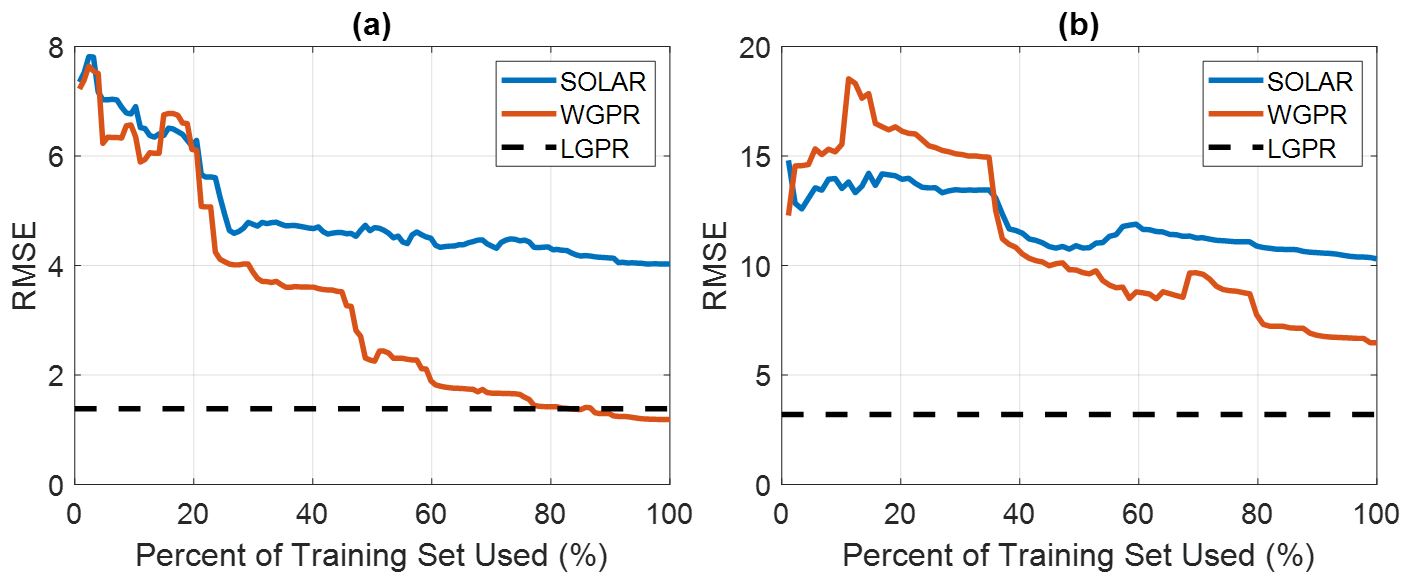}
  \caption{Evolution of the RMSE as more training batches are processed for Claytor Lake (a) and the Sarcos (b) datasets for SOLAR GP \cite{wilcox2020solar}, WGPR, and the offline benchmark LGPR \cite{nguyen2009model}. Observe that as more training examples are processed, we approach the performance of the offline method.\vspace{-6mm}}
\label{fig: offline comp}
\end{figure}

\subsection{Offline Benchmark} \label{ss: offline comp}
In various practical applications, there is no \textit{a priori} access to a dataset, characteristic of the entire state space, to learn the hyperparameters of the predictive model. To cope with these situations, we require that our method have the ability to iteratively update the hyperparameters of the predictive model as new training data is received. As local Gaussian process regression (LGPR) \cite{nguyen2009model} learns the hyperparameters on a characteristic subset of all training data prior to building the prediction model, we consider this an appropriate offline benchmark to assess the predictive performance of WGPR and SOLAR GP. Given that there are parameters unique to each method and LGPR employs the exact posterior \eqref{eq: post} for each model, we manually tune the parameters to achieve the best possible predictive performance (averaged across 5 trials) for all algorithms and compare them. 

We depict the results for Claytor Lake and the Sarcos dataset in Figure \ref{fig: offline comp} (a) and (b), respectively. Here, we see that as more training data is collected, the predictive performance of WGPR tends to LGPR and in both cases WGPR outperforms SOLAR GP. Moreover, for the Claytor Lake dataset, WGPR outperforms LGPR. This is consistent with the intuition that given the dataset is known to be nonstationary, we expect WGPR to outperform LGPR, as LGPR requires all models to have the same set of hyperparameters whereas WGPR does not. For all datasets, we collect the training time (seconds) and RMSE after processing the entire dataset in Tables \ref{tab: offline_times} and \ref{tab: offline_rmse}, respectively. We delineate LGPR from all other methods that iteratively learn hyperparameters using a dashed lined. We also included the online VFE method of Bui et al. \cite{bui2017streaming}, which we denote by OVFE. Note that entries in the table containing a ``-" indicate the algorithm failed to converge. By and large, we see that at the expense of additional computational time WGPR either outperforms or attains comparable performance with all other methods that iteratively learn the hyperparameters, and we achieve comparable performance with the offline benchmark. Note that LGPR exhibits the least computational time as all hyperparameter learning is only done once, prior to allocating samples to their respective models. The methods that learn hyperparameters online require more computation time, and as WGPR computes the Wasserstein metric over various candidate posterior distributions it incurs the greatest computational time.

\begin{table}[htbp]
  \centering
  \small
  \caption{Training time (s) of online methods and the offline benchmark LGPR}
    \resizebox{\columnwidth}{!}{\begin{tabular}{lccccc}
    \toprule
          & Claytor Lake & Sarcos & kin40k & Abalone & Pumadyn \\
    \midrule
    LGPR  \cite{nguyen2009model} & 39.484 & 347.857 & 31.221 & 13.221 & 38.138 \\\hdashline
    WGPR  & 210.053 & 870.358 & 202.184 & 62.912 & 120.836 \\
    SOLAR \cite{wilcox2020solar} & 131.417 & 477.143 & 119.403 & 40.126 & 76.353 \\
    OVFE \cite{bui2017streaming} & -     & -     & 115.902 & -     & 45.267 \\
    \end{tabular}}%
  \label{tab: offline_times}%
  \normalsize
  \vspace{-5mm}
\end{table}%

\begin{table}[htbp]
  \centering
  \small
  \caption{RMSE of online methods and the offline benchmark LGPR}
    \resizebox{\columnwidth}{!}{\begin{tabular}{lccccc}
    \toprule
          & Claytor Lake & Sarcos & kin40k & Abalone & Pumadyn \\
    \midrule
    LGPR \cite{nguyen2009model} & 1.381 & 3.195 & 0.231 & 2.07  & 1.079 \\ \hdashline
    WGPR  & 1.362 & 4.476 & 0.797 & 2.487 & 1.137 \\
    SOLAR \cite{wilcox2020solar} & 3.524 & 12.686 & 0.910 & 2.887 & 1.085 \\
    OVFE \cite{bui2017streaming}  & -     & -     & 0.996 & -     & 2.482 \\
    \end{tabular}}%
  \label{tab: offline_rmse}%
  \normalsize
  \vspace{-5mm}
\end{table}%

\section{Conclusion} \label{sec: Conclusion}
Our empirical results demonstrate that WGPR achieves comparable, if not superior, predictive performance to other methods that iteratively learn hyperparameters. 
Our new measure of similarity not only allows us to succinctly characterize a non-stationary spatial field, but more importantly offers the ability to prevent the pathology that can arise when using proximity-based similarity metrics: measurements from two regions of the spatial field, best characterized by different hyperparameters, are merged to form a single model because the regions were sufficiently close in proximity.
%

For WGPR, there are two user-specified parameters: the instantiation threshold $\epsilon$, and the number of inducing inputs $M$ in each Gaussian process model.
In practice, $\epsilon$ can be made ``large" to yield a map representation with fewer models if memory or transmitting data is a concern, as it is in communication-limited environments like the sub-sea domain \cite{keplerapproach}. However, this is at the potential cost of a loss in accuracy, as two models that may be best characterized by two distinct models may effectively be averaged into a single model. In contrast, if the size of the map representation is not a concern, $\epsilon$ can be made ``small'' in an effort to give rise to more models that may be more accurate. However it remains an open question of how to precisely define ``large'' and ``small'' as the similarity metric \eqref{eq: sim metric} is not generalizable across data sets. 

An interesting avenue of further research is to conduct isolation studies to understand the individual implications of each element: (i) Wasserstein-splitting, (ii) prediction of the nearest model, and (iii) the expectation propagation steps. Additionally, studies to better understand how computing the similarity metric over the $\hat{J} \leq J$ nearest models impacts performance are of interest. Lastly, we seek principled methods to determine the number of pseudo-points in each local model to ensure a certain quality of approximation.




\bibliographystyle{./IEEEtran}
\bibliography{./IEEEabrv,./mybibfile}

\end{document}